\documentclass[]{fairmeta}

\usepackage{tikz}
\usetikzlibrary{positioning, shapes.geometric, arrows.meta, fit,calc, matrix, arrows.meta}
\usepackage{fontawesome}
\usepackage{contour}
\usepackage{xspace}

\newcommand{\gc}[1]{\textcolor{gray!70}{#1}}
\DeclareMathOperator{\largesymbol}{\Large}

\definecolor{colorOne}{RGB}{179, 226, 205}
\definecolor{colorTwo}{RGB}{253, 205, 172}
\definecolor{colorThree}{RGB}{203, 213, 232}
\definecolor{colorFour}{RGB}{244,202,228}
\definecolor{colorFive}{RGB}{230,245,201}

\title{Efficient Speculative Decoding for Llama at Scale: Challenges and Solutions}

\author[\largesymbol{\heartsuit}]{GenAI and Infra Teams at Meta.}

\affiliation[\largesymbol{\heartsuit}]{A detailed author list can be found in the Contributions section of this paper.}

\abstract{
Speculative decoding is a standard method for accelerating the inference speed of large language models. However, scaling it for production environments poses several engineering challenges, including efficiently implementing different operations (e.g., tree attention and multi-round speculative decoding) on GPU. In this paper, we detail the training and inference optimization techniques that we have implemented to enable EAGLE-based speculative decoding at a production scale for Llama models. With these changes, we achieve a new state-of-the-art inference latency for Llama models. For example, Llama4 Maverick decodes at a speed of about 4 ms per token (with a batch size of one) on 8 NVIDIA H100 GPUs, which is 10\% faster than the previously best known method. Furthermore, for EAGLE-based speculative decoding, our optimizations enable us to achieve a speed-up for large batch sizes between $1.4\times$ and $2.0\times$ at production scale.
}

\date{\today}
\correspondence{Sachin Mehta at \email{sacmehta@meta.com}}

\begin{document}

\maketitle

\section{Introduction}
\label{section:intro}
Attention-based transformers of \citet{vaswani2017attention} are widely used across different areas of machine learning and artificial intelligence, including natural language processing, computer vision, and speech processing. Specifically, transformer-based large language models (LLMs) have been scaled to billions of parameters, and are trained on trillions of tokens to produce high-quality models \citep[e.g.,][]{achiam2023gpt, grattafiori2024llama, llama4, team2023gemini}. Because of their auto-regressive nature and large size, the decoding speed fo these models' is very slow, posing significant challenges when deployed in production environments where they must handle a large volume of requests with varying input and output lengths.

Several methods, including FlashAttention \citep{dao2022flashattention,dao2023flashattention2}, memory-efficient attention \cite{rabe2021self},  fully-sharded data parallel \citep{FairScale2021}, and disaggregated inference \citep{zhong2024distserve,qin2407mooncake}, have been proposed to improve the training and inference speed of transformer-based models. Complementary to these methods, speculative decoding \citep{leviathan2023fast,chen2023accelerating} has emerged as a promising technique for accelerating the inference speed of LLMs. Briefly, speculative decoding involves predicting multiple tokens using a smaller model (aka, draft model) auto-regressively, which are validated in a single step using an LLM. This reduces the number of calls to the auto-regressive LLM, improving decoding speed. However, this approach requires significantly more floating point operations (FLOPs) than the non-speculative decoding model because multiple tokens need to be validated by the LLM as opposed to single token decoding in case of a non-speculative decoding setup. Several speculative decoding methods have been proposed \citep[e.g.,][]{cai2024medusa,miao2024specinfer,li2024eagle}. However, most of these methods are benchmarked at small scale settings (e.g., batch size of one) to demonstrate the speed-up over non-speculative decoding method, likely because of significant engineering challenges in scaling these methods to production environments that require handling a large volume of dynamic requests efficiently. For example, when the batch size is increased from 2 to 48, the speed-up (measured using vLLM) of EAGLE-based speculative decoding compared to non-speculative decoding drops from $1.3\times$ to $0.7\times$ (see Table 5 in v3 of \citet{li2025eagle}). Similar behaviour was also observed with SGLang \citep{sglang} (see Table 5 in \citep{li2025eagle}). 

\begin{figure}[t!]
    \centering
    \begin{subfigure}[b]{0.48\columnwidth}
        \centering
        \includegraphics[width=\columnwidth]{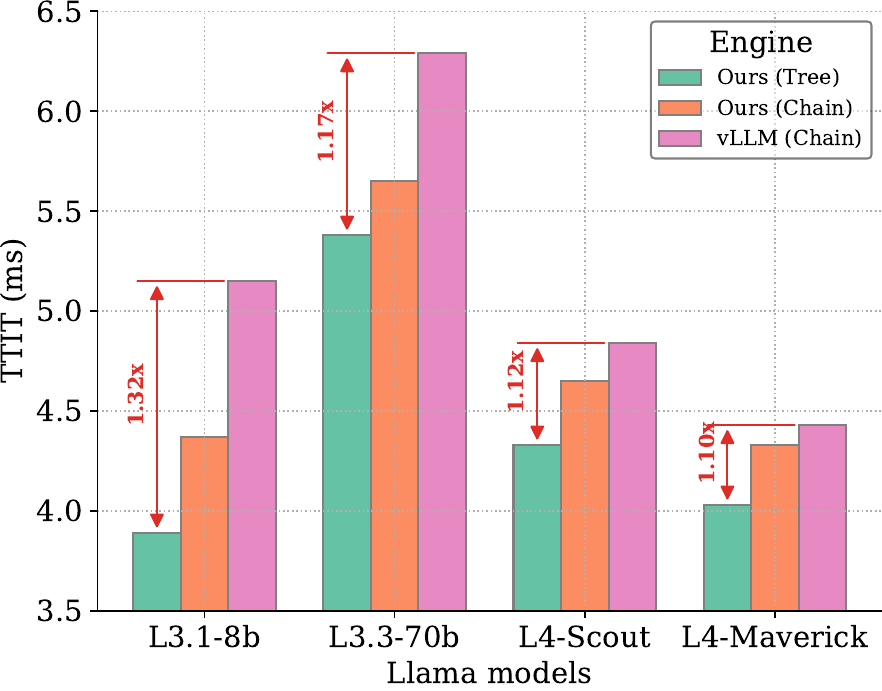}
        \caption{}
        \label{fig:compare_vllm}
    \end{subfigure}
    \hfill
    \begin{subfigure}[b]{0.48\columnwidth}
        \centering
        \includegraphics[width=\columnwidth]{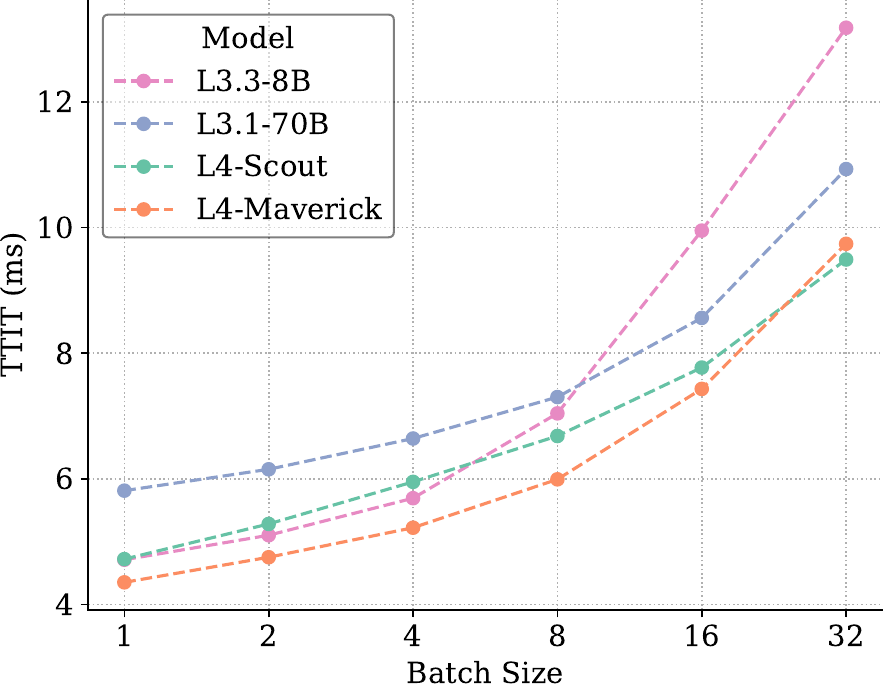}
        \caption{}
        \label{fig:batch_size_vs_ttit}
    \end{subfigure}
    \vfill
    \begin{subfigure}[b]{0.75\columnwidth}
        \centering
        \includegraphics[width=\columnwidth]{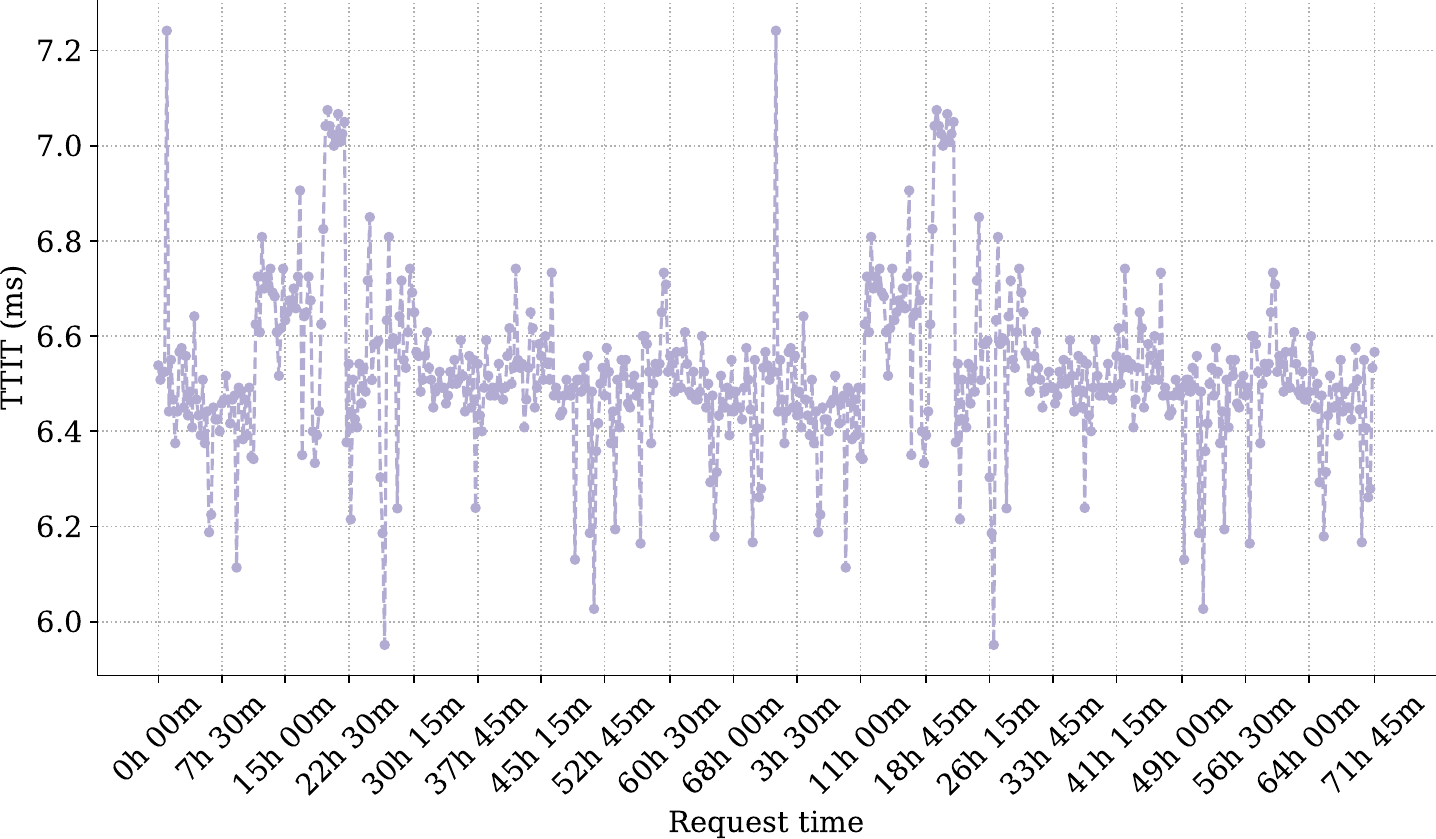}
        \caption{}
        \label{fig:l3_prod}
    \end{subfigure}
    
    \caption{Performance evaluation of our EAGLE-based speculative decoding system. (a) TTIT comparison of our system with vLLM at a batch size of one. (b) TTIT comparison of different LLaMA models within our system at various batch sizes and a context length of 8k. (c) TTIT for online user requests using LLaMA 3.3 70B over a 3-day period. \gc{In (b), we do not compare with vLLM because of significant gaps in TTIT between our system and vLLM. This is likely because vLLM stacks might not be optimized for large batch sizes, as also observed in other speculative decoding works \citep[e.g.,][]{li2025eagle}.
    }}
    \label{fig:perf_eval}
\end{figure}

In this paper, we detail the training (see \Cref{sec:train_optim}) and inference (see \Cref{sec:infer_optim}) optimization techniques we have implemented to enable EAGLE-based speculative decoding for Llama models at a production scale, addressing the challenges associated with deploying these models in real-world applications. \Cref{fig:perf_eval} shows that, with our optimizations, the decoding speed (with a batch size of one) of Llama models with EAGLE on 8 NVIDIA H100 GPUs improves by about 10-30\% as compared to widely used open-source library vLLM. Moreover, at large batch sizes, our optimizations for EAGLE-based speculative decoding further enable a speed-up between $1.4\times$ and $2\times$. We note that the proposed optimizations are complementary to open-source libraries \citep[e.g.,][]{kwon2023efficient, sglang} and can be easily integrated to further improve the inference of supported speculative decoding methods.

\section{Training optimizations}
\label{sec:train_optim}
We introduce three changes for EAGLE-based speculative decoding: (a) online distillation (see \Cref{ssec:online_distillation}), (b) longer training (see \Cref{ssec:longer_training}), and (c) multi-layer dense draft model (see \Cref{ssec:dense_draft_model}). With these changes, we trained draft models for four different Llama models that differs in several aspects, including architecture and model size. The results are then presented in \Cref{ssec:tpc_results_llama}. 

\subsection{Online distillation} 
\label{ssec:online_distillation}

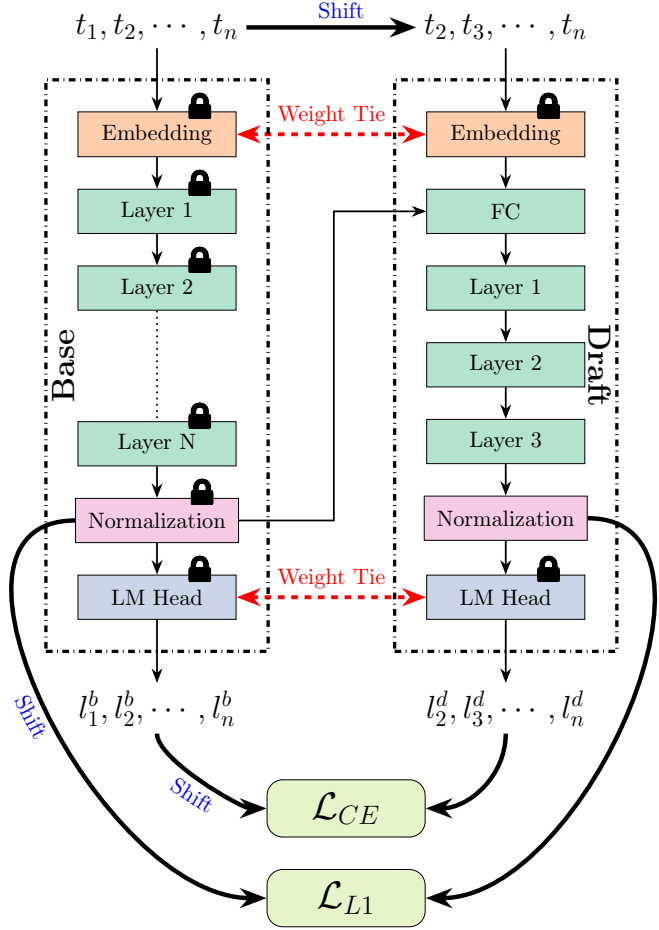
\begin{figure}[t!]
    \centering
    \resizebox{0.7\columnwidth}{!}{
        \newcommand{\modelDiagram}{
    \begin{tikzpicture}[
        transformer/.style={rectangle, draw, minimum width=2.5cm, minimum height=0.7cm, align=center, fill=colorOne},
        frozen/.style={label={[label distance=-0.9cm]north east:\contour{black}{\textcolor{black}{\Large \faLock}}}},
        layer/.style={rectangle, draw, minimum width=2.5cm, minimum height=0.7cm, align=center, fill=colorTwo, inner sep=0.2cm},
        arrow/.style={-Stealth, thick},
        sharedArr/.style={Stealth-Stealth, line width=1.75pt, dashed, red},
        node distance=1.5cm
    ]
    \node (input) [layer, frozen] {Embedding};
    \node (base1) [transformer, frozen, below=0.5cm of input] {Layer 1};
    \node (base2) [transformer, frozen, below=0.5cm of base1] {Layer 2};
    \node (baseN) [transformer, frozen, below=1.75cm of base2] {Layer N};
    \node (norm) [layer, frozen, below=0.5cm of baseN, fill=colorFour] {Normalization};
    \node (class) [layer, frozen, below=0.5cm of norm, fill=colorThree] {LM Head};
    \node (inputD) [layer, frozen, right=3cm of input] {Embedding};
    \node (fc) [transformer, below=0.5cm of inputD] {FC};
    \node (base1D) [transformer, below=0.5cm of fc] {Layer 1};
    \node (base2D) [transformer, below=0.5cm of base1D] {Layer 2};
    \node (baseMD) [transformer, below=0.5cm of base2D] {Layer 3};
    \node (normD) [layer, below=0.5cm of baseMD, fill=colorFour] {Normalization};
    \node (classD) [layer, frozen, right=3cm of class, fill=colorThree] {LM Head};

    \draw[arrow] (input) -- (base1);
    \draw[arrow] (base1) -- (base2);
    \draw[thick, dotted] (base2) -- (baseN);
    \draw[arrow] (baseN) -- (norm);
    \draw[arrow] (norm) -- (class);
    \draw[sharedArr] (input.east) -- ++(0.5,0) node[pos=3, above] {Weight Tie} |- (inputD.west);
    \draw[arrow] (norm.east) -- ++(1.5,0) |- (fc.west);
    \draw[arrow] (inputD) -- (fc);
    \draw[arrow] (fc) -- (base1D);
    \draw[arrow] (base1D) -- (base2D);
    \draw[arrow] (base2D) -- (baseMD);
    \draw[arrow] (baseMD) -- (normD);
    \draw[arrow] (normD) -- (classD);
    \draw[sharedArr] (class.east) -- ++(0.5,0) node[pos=3, above] {Weight Tie} |- (classD.west);
    \node[above=1cm of input] (inpTokens) {\Large $t_1, t_2, \cdots, t_n$};
    \node[below=1cm of class] (logitsT) {\Large $l^b_1, l^b_2, \cdots, l^b_n$};
    \draw[arrow] (inpTokens) -- (input);
    \draw[arrow] (class) -- (logitsT);
    \node[above=1cm of inputD] (inpTokensD) {\Large $t_2, t_3, \cdots, t_n$};
    \node[below=1cm of classD] (logitsTD) {\Large $l^d_2, l^d_3,\cdots, l^d_n$};
    \draw[arrow] (inpTokensD) -- (inputD);
    \draw[arrow] (classD) -- (logitsTD);
    \draw[arrow, line width=2pt] (inpTokens.east) -- ++(0.5,0) node[pos=3, above, text=blue] {Shift} |- (inpTokensD.west);
    \node (dummy) [right=1.5cm of logitsT] {};
    \node (CE) [layer, below=1cm of dummy, fill=colorFive, rounded corners=0.25cm,] {\LARGE $\mathcal{L}_{CE}$};

    \draw[arrow, line width=2pt] (logitsT.south) to[out=-90, in=180, looseness=0.5] node[midway, below, sloped, text=blue] {Shift} (CE.west);
    
    \draw[arrow, line width=2pt] (logitsTD.south) to[out=-90, in=0] (CE.east);

    \node (L1) [layer, below=0.5cm of CE, fill=colorFive, rounded corners=0.25cm,] {\LARGE $\mathcal{L}_{L1}$};
    \draw[arrow, line width=2pt] (normD.east) to[out=0, in=0] (L1.east);
    \draw[arrow, line width=2pt] (norm.west) to[out=180, in=180, looseness=1] node[midway, below, sloped, text=blue] {Shift} (L1.west);

    \node[fit=(input) (class), draw, dashdotted, inner sep=0.5cm, line width=1.5pt] (main model) {};
    \node[rotate=90] at ($(main model.west) + (0.3, 0)$) {\Large \bfseries Base};

    \node[fit=(inputD) (classD), draw, dashdotted, inner sep=0.5cm, line width=1.5pt] (draft model) {};
    \node[rotate=-90] at ($(draft model.east) + (-0.3, 0)$) {\Large \bfseries Draft};
    \end{tikzpicture}
}\modelDiagram
    }
    \caption{Overview of online distillation that we used to train EAGLE speculative decoding method for Llama-3 and Llama-4 models. Here, \faLock~\xspace represents frozen layers.}
    \label{fig:eagle_online_distill}
\end{figure}

An overview of our implementation of EAGLE speculative decoding is given in \Cref{fig:eagle_online_distill}. Specifically, it uses a base model 
$\mathcal{B}$ and a draft model $\mathcal{D}$, both of which are auto-regressive. The base model consists of $N$ sequential transformer layers, while the draft model is light-weight with $M$ layers, where $M <<< N$. During training, the base model $\mathcal{B}$ takes $n$ tokens $\mathbf{t} = \{t_1, t_2, \cdots, t_n\}$ as input. It produces  hidden states $\mathbf{h}^b = \{h^b_1, h^b_2, \cdots, h^b_n\}$ (i.e., output from the normalization layer before LM head) and logits $\mathbf{l^b} = \{l^b_1, l^b_2, \cdots, l^b_n\}$ (i.e., output from the LM head before softmax). For the draft model, the input tokens are shifted to the right by one, resulting in $\mathbf{\hat{t}} = \{t_2, t_3, \cdots, t_n\}$. These tokens are fed into the embedding layer. Unlike the base model, the draft model includes a fully-connected layer between embedding and the first transformer layer. The layer takes the token embeddings of $\mathbf{\hat{t}}$ and hidden states $\mathbf{h}$ from the base model as inputs, and the resulting output is fed to the rest of the draft model to produce the hidden states $\mathbf{h}^d = \{h^d_2, h^d_3, \cdots, h^b_n\}$ and logits $\mathbf{l^d} = \{l^d_2, l^b_3, \cdots, l^b_n\}$.

To train the draft model, we minimize the loss between (a) hidden states and (b) logits of the base $\mathcal{B}$ and the draft $\mathcal{D}$ models. Specifically, we compute the smooth L1 loss between the right shifted hidden states of the base model and the hidden states of the draft model, as $\mathcal{L}_{L1} = \text{l1\_loss}(h^b_{2:n}, h^d_{2:n})$. Similarly, we compute the cross entropy loss between shifted logits of the base model and those of the draft model, as $\mathcal{L}_{CE} = \text{ce\_loss}(l^b_{2:n}, l^d_{2:n})$. The final loss is a weighted sum between $\mathcal{L}_{L1}$ and $\mathcal{L}_{CE}$, as
\begin{equation}
    \mathcal{L} = \lambda_{CE} \cdot \mathcal{L}_{CE} + \lambda_{L1} \cdot \mathcal{L}_{L1}
\end{equation} 
where $\lambda_{CE}$ and $\lambda_{L1}$ are the coefficients to control the contributions of cross entropy and L1 loss, respectively. In our experiments, we use $\lambda_{CE}=0.1$ and $\lambda_{L1}=1.0$. 

\subsection{Longer training} 
\label{ssec:longer_training}

We train draft models for the following four base models belonging to Llama3 \citep{grattafiori2024llama} and Llama4 \citep{llama4} families: (1) Llama3.1 8B, (2) Llama3.1 70B, (3) Llama4 Scout (total and active parameters are about 109B and 17B respectively), and (4) Llama4 Maverick (total and active parameters are about 400B and 17B respectively). The models in Llama3 family uses dense feed-forward network (FFN) layers while Llama4 models uses mixture-of-experts (MoE; \citep{shazeer2017outrageously}) layers. We use the same supervised fine-tuning (SFT) dataset that was used to train the models for a total of 48k iterations, with 2M tokens per iteration.  We optimize the draft model using Adam \cite{kingma2014adam}, with a constant learning rate of 0.0002 and a weight decay of 0.1. To evaluate the quality of the draft model, we measured the number of accepted tokens per drafting-validation stage (or tokens per call; TPC) on two benchmarks: (a) MT-Bench \citep{zheng2023judging}, a public benchmark that is widely used for measuring speculative decoding performance, and (2) a private internal benchmark, that contains a diverse, multi-lingual and harder samples. We used a batch size of one and chain-like draft with speculation length of three for TPC evaluation. A higher value of TPC is desirable.

\begin{figure}[t!]
    \centering
    \begin{subfigure}[b]{0.4\columnwidth}
        \includegraphics[width=\columnwidth]{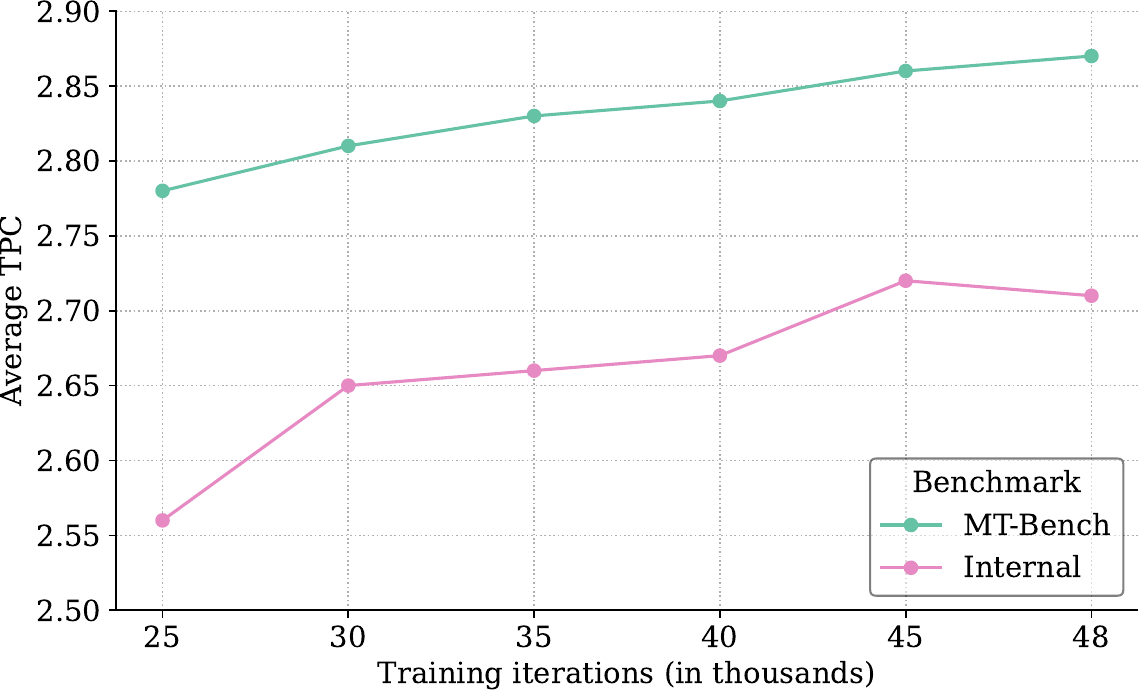}
        \caption{}
        \label{fig:longer_training}
    \end{subfigure}
    \hfill
    \begin{subfigure}[b]{0.55\columnwidth}
        \resizebox{0.9\columnwidth}{!}{
            \begin{tabular}{ccccccc}
                \toprule[1.5pt]
                \multirow{2}{*}{\textbf{\# Layers}} & \multicolumn{2}{c}{\textbf{Model configurations}} && \multicolumn{2}{c}{\textbf{TPC} $\bm{\uparrow}$} \\
                \cmidrule[1.25pt]{2-3}
                \cmidrule[1.25pt]{5-6}
                & \textbf{iRoPE} & \textbf{FFN Type} && \textbf{MT-Bench} & \textbf{Internal} \\
                \midrule[1.25pt]
                \multirow{4}{*}{2} & \multirow{2}{*}{Yes} & MoE && 2.75 & 2.55 \\
                 & & Dense && 2.79 & 2.58 \\
                 \cmidrule[1pt]{2-6}
                  & \multirow{2}{*}{No} & MoE && 2.75 & 2.55 \\
                  & & Dense && 2.80 & 2.58 \\
                \midrule[1pt]
                3 & No & Dense && \textbf{2.87} & \textbf{2.71} \\
                \bottomrule[1.5pt]
                \end{tabular}
        }
        \caption{}
        \label{fig:spec_decoding_results_scout}
    \end{subfigure}
    \caption{Ablation with Llama4 Scout. (a) Effect of training duration on the Llama4 Scout's TPC on two benchmarks, MT-Bench and internal benchmarks. (b) Effect of different draft model design choices for Llama4 Scout base model. Here, TPC is  measured with chain-like draft, temperature=0, top-p=0.9, and speculation length of three. iRoPE denotes interleaved RoPE.}
    \label{fig:ablations_scout}
\end{figure}

\Cref{fig:longer_training} shows the effect of longer draft training using Llama4 Scout. We see that longer training helps improve the average tokens accepted per drafting step or tokens accepted per call (TPC) on both benchmarks. Note that we observe a similar trend in other models as well.

\subsection{Multi-layer dense draft models} 
\label{ssec:dense_draft_model}
For the draft model design, we considered transformer blocks with dense and MoE FFN. We also studied the relationship between the number of transformer blocks and the quality metrics. The results are shown in \Cref{fig:spec_decoding_results_scout}. We make following observations:

\begin{itemize}
    \item \textbf{Dense vs. MoE FFN:} Our results indicate that draft models with dense FFN achieve comparable TPC to MoE, while utilizing substantially fewer parameters. For example, the draft model for Llama4 Maverick with dense FFN layers has about $10 \times$ fewer total parameters as compared to MoE. Therefore, we used dense FFN layers in our draft models.
    \item \textbf{Effect of iRoPE:} Llama4 introduced interleaved ROPE (iRoPE). We did not observe any significant differences in TPC or end-to-end latency with and without iRoPE. Therefore, iRoPE was not used in the training of our final draft models.
    \item \textbf{Effect of number of layers:} We varied the number of layers from 1 to 3 and observed a notable improvement in TPC, from 2.63 to 2.87. However, further increasing the model's depth beyond 3 layers did not yield significant gains in TPC. Therefore, we choose a 3-layer dense draft model as our final candidate for speculative decoding.
\end{itemize}

\subsection{Results} 
\label{ssec:tpc_results_llama}

Results for different Llama3 and Llama4 models on the MT-bench benchmark are shown in \Cref{tab:tpc_llama_models}. The TPC values vary across models, with Llama3.3 70B achieving the highest value of 2.94 and Llama4-Maverick achieving the lowest value of 2.75, likely due to differences in model capacity and architecture. \gc{As a reference, \Cref{tab:tpc_llama_models} also includes the original results for the 8B and 70B models with both EAGLE and EAGLE3 \citep{li2025eagle}, measured using vLLM \citep{kwon2023efficient} at different speculation lengths. With the proposed changes, EAGLE achieves similar or better TPC than EAGLE3, highlighting the effectiveness of the introduced training optimizations. Interestingly, EAGLE with the proposed changes and a speculation length of three outperforms the EAGLE from \citet{li2024eagle} even at a speculation length of seven.}

\begin{table}[t!]
    \centering
    \begin{tabular}{lcc}
        \toprule[1.5pt]
        \textbf{Model} & \textbf{Speculation length} & \textbf{TPC} $\bm{\uparrow}$ \\
        \midrule[1pt]
        \gc{Llama3.1-8B w/ EAGLE} & \gc{3/5/7} & \gc{2.29/2.44/2.47} \\
        \gc{Llama3.1-8B w/ EAGLE3} & \gc{3/5/7} & \gc{2.80/3.32/3.57} \\
        \midrule[1pt]
        \gc{Llama3.1-8B  w/ EAGLE} & \gc{3/5/7} & \gc{2.12/2.24/2.27}\\
        \gc{Llama3.3-70B  w/ EAGLE3} & \gc{3/5/7} & \gc{2.64/3.03/3.20}\\
        \midrule[1pt]
        Llama3.1-8B w/ EAGLE (Ours) & 3 & 2.78 \\
        Llama3.3-70B w/ EAGLE (Ours) & 3 & 2.94 \\
        Llama4-Scout w/ EAGLE (Ours) & 3 & 2.87 \\
        Llama4-Maverick w/ EAGLE (Ours) & 3 & 2.75 \\
        \bottomrule[1.5pt]
    \end{tabular}
    \caption{TPC results for different Llama models on the MT-Bench benchmark. Here, TPC is  measured with chain-like draft, temperature=0, and top-p=0.9.}
    \label{tab:tpc_llama_models}
\end{table}

\section{Inference optimizations}
\label{sec:infer_optim}
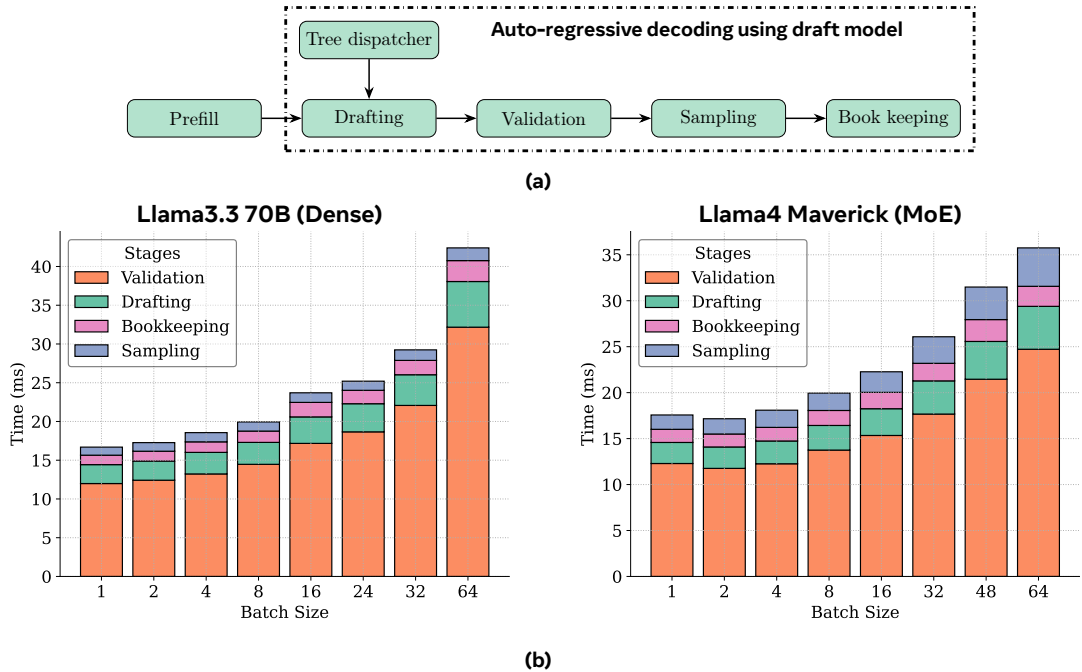
\begin{figure}[t!]
    \centering
    \begin{subfigure}[b]{\columnwidth}
        \centering
        \resizebox{0.7\columnwidth}{!}{
            \newcommand{\inferenceFlow}{
    \begin{tikzpicture}[
        transformer/.style={rectangle, draw, minimum width=2.5cm, minimum height=0.7cm, align=center, fill=colorOne, rounded corners},
        layer/.style={rectangle, draw, minimum width=2.5cm, minimum height=0.7cm, align=center, fill=colorTwo, inner sep=0.2cm},
        arrow/.style={-Stealth, thick},
        sharedArr/.style={Stealth-Stealth, line width=1.75pt, dashed, red},
        node distance=1.5cm
    ]
        \node (treeDispatch) [transformer] {Tree dispatcher};

        \node (drafter) [transformer, below=0.75cm of treeDispatch] {Drafting};

        \node (prefill) [transformer, left=0.75cm of drafter] 
        {Prefill};

        \node (validation) [transformer, right=0.75cm of drafter] {Validation};

        \node (sampling) [transformer, right=0.75cm of validation] {Sampling};

        \node (bookkeeping) [transformer, right=0.75cm of sampling] {Book keeping};

        \draw[arrow] (prefill) -- (drafter);
        \draw[arrow] (treeDispatch) -- (drafter);
        \draw[arrow] (drafter) -- (validation);
        \draw[arrow] (validation) -- (sampling);
        \draw[arrow] (sampling) -- (bookkeeping);

        \node[fit=(treeDispatch) (bookkeeping), draw, dashdotted, inner sep=0.25cm, line width=1.5pt] (draftStageFull) {};

        \node (temp) at (draftStageFull.north) [anchor=north, yshift=-0.15cm, xshift=1.25cm] {\large \textbf{Auto-regressive decoding using draft model}};
    \end{tikzpicture}
}\inferenceFlow
        }
        \caption{}
        \label{fig:inference_overview}
    \end{subfigure}
    \vfill
    \begin{subfigure}[b]{\columnwidth}
        \centering
        \begin{tabular}{ccc}
            \textbf{Llama3.3 70B (Dense)} & & \textbf{Llama4 Maverick (MoE)} \\
             \includegraphics[width=0.4\columnwidth]{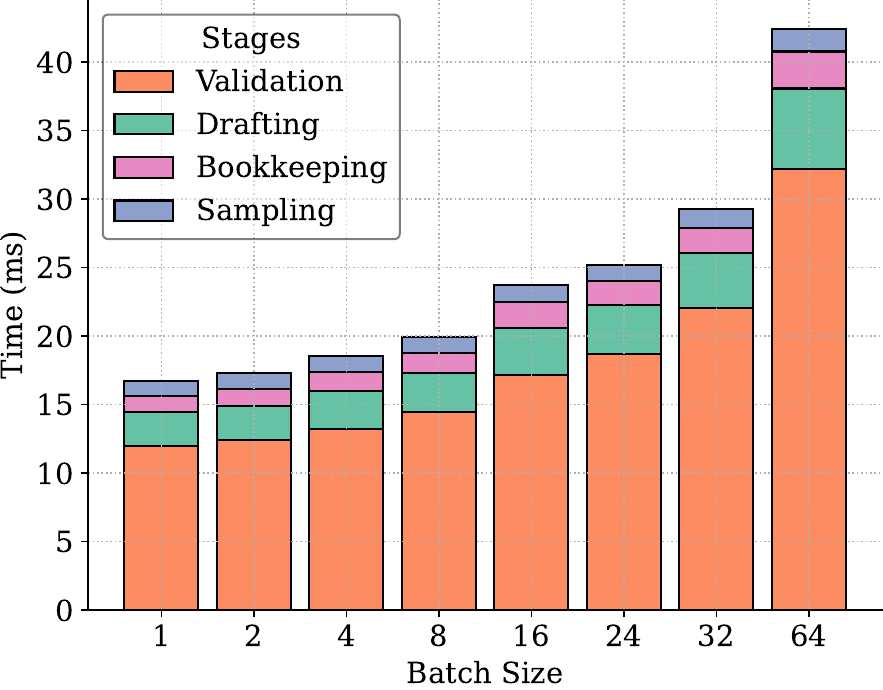}
             & \hspace{3pt} &
             \includegraphics[width=0.4\columnwidth]{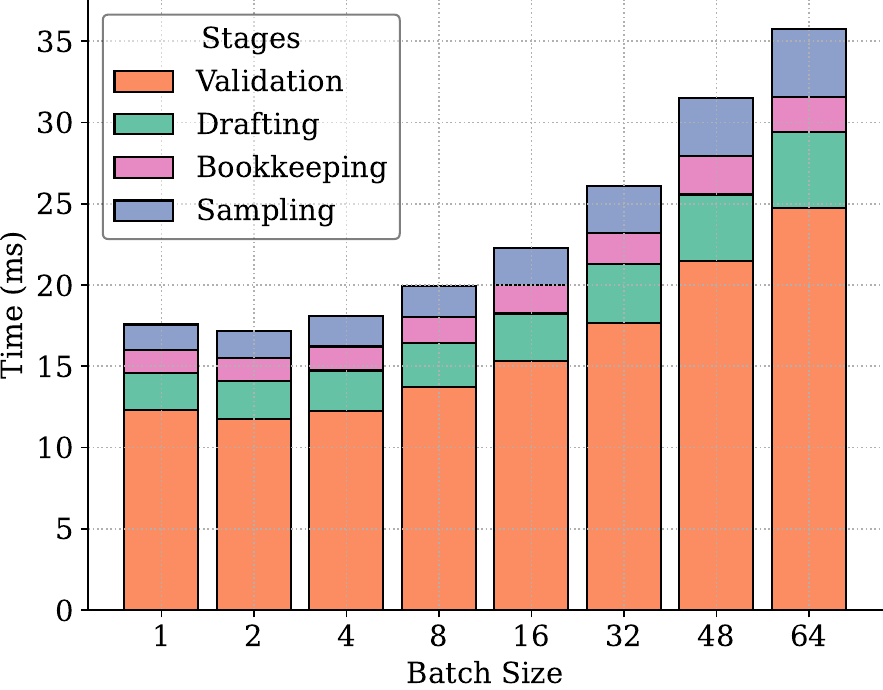}
             \\
        \end{tabular}
    \caption{}
    \end{subfigure}
    \caption{\textbf{Inference Workflow}. (a) Flow diagram illustrating EAGLE speculative decoding in our production environment. (b) Example showing a breakdown of the decoding step latency for each stage of the inference workflow at varying batch sizes, measured for Llama3.3 70B and Llama4 Maverick on an NVIDIA H100 GPU. Note that, in this example, three tokens are generated auto-regressively from the draft model and validated by the base model in a single decoding step. Because the decoding step generates multiple tokens, it should not be confused with TTIT, which measures the speed of decoding a single token.}
    \label{fig:inference_flow}
\end{figure}

Our EAGLE speculative decoding inference process, shown in \Cref{fig:inference_flow}, is organized into the following stages: (1) \textit{Prefilling:} This stage involves executing the base model prefill on input prompt tokens, followed by the draft model prefill on the input tokens and hidden states produced by the base model. (2) \textit{Tree Dispatcher:} In this stage, an optimal static tree structure is selected for the current batch. (3) \textit{Drafting:} The EAGLE drafter takes the previous token and its hidden state from the base model, running draft model auto-regressively to construct a tree of draft tokens. (4) \textit{Validation:} Here, the draft tokens are verified using the base model. The tree-attention-based inference enables efficient one-shot validation of the entire tree of draft tokens. (5) \textit{Sampling:} This stage involves deciding whether to accept some or all of the proposed tokens. Following EAGLE, we use multi-round speculative sampling which  preserves the output probability distribution of the original LLM model. (6) \textit{Bookkeeping:} EAGLE speculative decoding requires caching hidden states for the next round of drafting. Therefore, during bookkeeping, the KV cache and hidden states are rewounded to the appropriate position based on the accepted length.

\begin{figure}[t!]
    \centering
    \includegraphics[width=0.7\columnwidth]{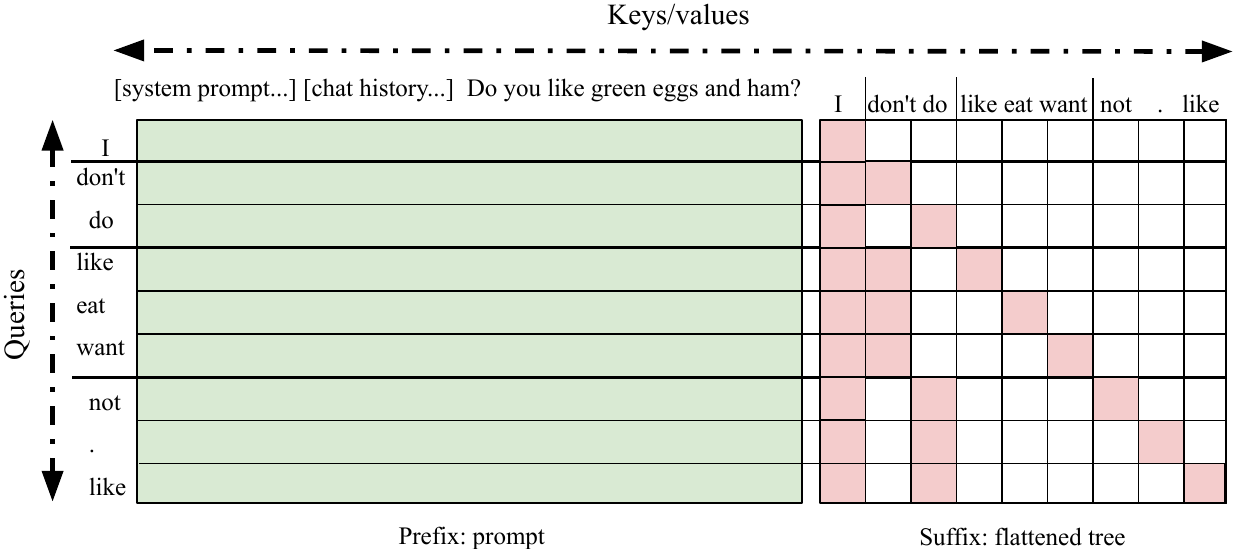}
    \caption{An illustration of optimized tree attention. Unlike the standard and unoptimized tree attention (see Figure 4 in \citet{miao2024specinfer}), we split the attention operation into prefix and suffix for efficient inference. }
    \label{fig:tree_attention}
\end{figure}

\subsection{Tree attention}
\label{ssec:tree_attention}

Tree attention \citep{miao2024specinfer} is an important technique in EAGLE speculative decoding, and is used in both drafting and verification stages. During each decoding round, one path in the tree is partially accepted based on an acceptance criterion, which utilizes validation logits generated by the target model at each node. A naive approach to compute these logits would involve unrolling all possible paths and performing a prefill computation with an effective batch size equal to the product of the batch size and the number of paths. However, a more efficient method would be to flatten all draft tokens into a single sequence. This approach introduces a challenge when computing attention, as the flattened tree tokens are not in the natural sequential order, making it hard to apply standard causal attention mask. 

A custom tree mask can be applied by providing an explicit mask tensor. However, tree attention with an explicit mask is slow due to the large shape of the mask, which scales with the query and context (keys/values) length. Therefore, we split the attention computation into two parts: (a) attention between the query and context (prefix), and (b) attention within the query itself (suffix), as shown in \Cref{fig:tree_attention}. The former is computationally expensive but mask-free, whereas the latter requires a tree mask but is relatively small. This two pass approach allows us to effectively prepare and extract query/key/values for speculated tokens needed for suffix calculations without any performance overhead. These two attention computations are then aggregated using \texttt{merge\textunderscore attentions}, a simple method which computes full attention based on partial attention outputs computed on two disjoint chunks of KV context \citep{juravsky2024hydragen}. We have used this method to implement efficient tree attention in xFormers \citep{xFormers2022}. It's worth mentioning that our optimized tree attention code in xFormers can be used seamlessly with other tree-based speculative decoding methods, such as SpecInfer \citep{miao2024specinfer} and Medusa \citep{cai2024medusa}, to further improve their inference efficiency.

\subsection{Multi-round speculative sampling} 
Multi-round speculative sampling (MSS) in EAGLE extends the standard speculative sampling method of \citet{leviathan2023fast, chen2023accelerating}, specifically adapting it for tree-like drafts. Naive implementation of MSS can introduce significant overhead in production decoding environments due to the nested loops over tree depth and the need to launch numerous small kernels. To address this, we implemented the following optimizations:
\begin{itemize}
    \item \textit{\bfseries PyTorch-2 compilation:} While most operations in MSS are not inherently computationally intensive, the nested loop over tree depth and the children of each node can result in significant CPU overhead, particularly in large-scale production environments. To mitigate this, we compiled the code using PyTorch-2, achieving a $1.5\times$ speedup. A key consideration was handling the dynamic batch dimension for variable incoming traffic. Naively applying \texttt{torch.compile} could lead to recompilation for each batch size, potentially causing latency spikes in production. To address this, we designated the batch dimension as dynamic in every input tensor. With this change, recompilations are limited to just two cases: batch size of one and batch sizes greater than one. Moreover, if a service receives warm-up traffic at startup, and both scenarios are covered, the compiler cache is populated by \texttt{torch.compile} during the warm-up phase. This cache is then reused during real traffic, helping to avoid latency spikes.
    \item \textit{\bfseries Parallelisation across tensor parallel ranks:} Despite the GPU-efficient implementation and PyTorch 2 compilation, sampling becomes a noticeable bottleneck in decoding performance at large batch sizes. One primary reason is the application of the top-p mask or nucleus sampling \citep{DBLP:conf/iclr/HoltzmanBDFC20}, which involves inherently ``serial" operations that are challenging to accelerate on the GPU. Observing that (a) sampling is embarrassingly parallel over the batch dimension and (b) all tensor parallel (TP) ranks should receive the same sampling result, we parallelized the sampling across TP ranks by padding the batch size to a multiple of the total number of GPUs (aka, world size). However, with this simple solution, special care must be taken to synchronize random number generators across ranks. Indeed, during sampling, each rank generates random tensors with sizes proportional to the batch size. If different ranks process different local batch sizes, such as when the last rank handles the remainder of the global batch size divided by the world size, the random number generators may diverge, potentially causing system hangs in subsequent iterations. To prevent this, we generate a full-sized random tensor on each rank and have each rank take a different slice of this tensor.
    \item \textit{\bfseries Greedy draft decoding:} Different methods for sampling draft tokens are known for tree-shaped drafts, including multinomial (with and without replacement) and greedy sampling \citep{miao2024specinfer, jeon2024recursive}. Specifically, for greedy decoding, we consider a node with $k$ children, and place $k$ tokens with the highest probabilities into these child nodes. In our experiments, we  observed slightly higher TPC from greedy sampling compared to other methods. Moreover, the computationally expensive top-p mask is a no-op for draft probabilities with greedy decoding, and allows us to skip top-p mask at the drafting stage (while still applying it to target logits at the validation stage). This  reduces the computational overhead from drafting and further helps in improving inference efficiency.
\end{itemize}

\begin{figure}[t!]
    \centering
    \begin{subfigure}[b]{\columnwidth}
        \includegraphics[width=\columnwidth]{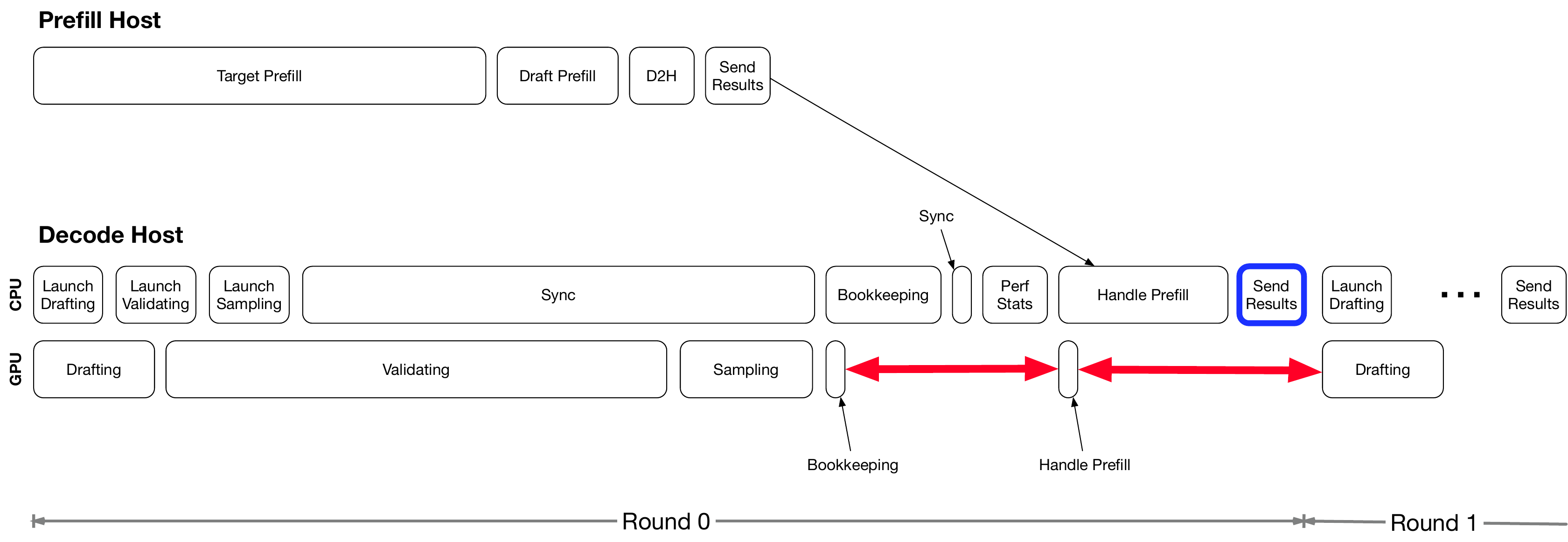}
        \caption{Before latency hiding optimizations}
        \label{fig:hide_cpu_ops_trace_before}
    \end{subfigure}
    \vfill
    \begin{subfigure}[b]{\columnwidth}
            \includegraphics[width=\columnwidth]{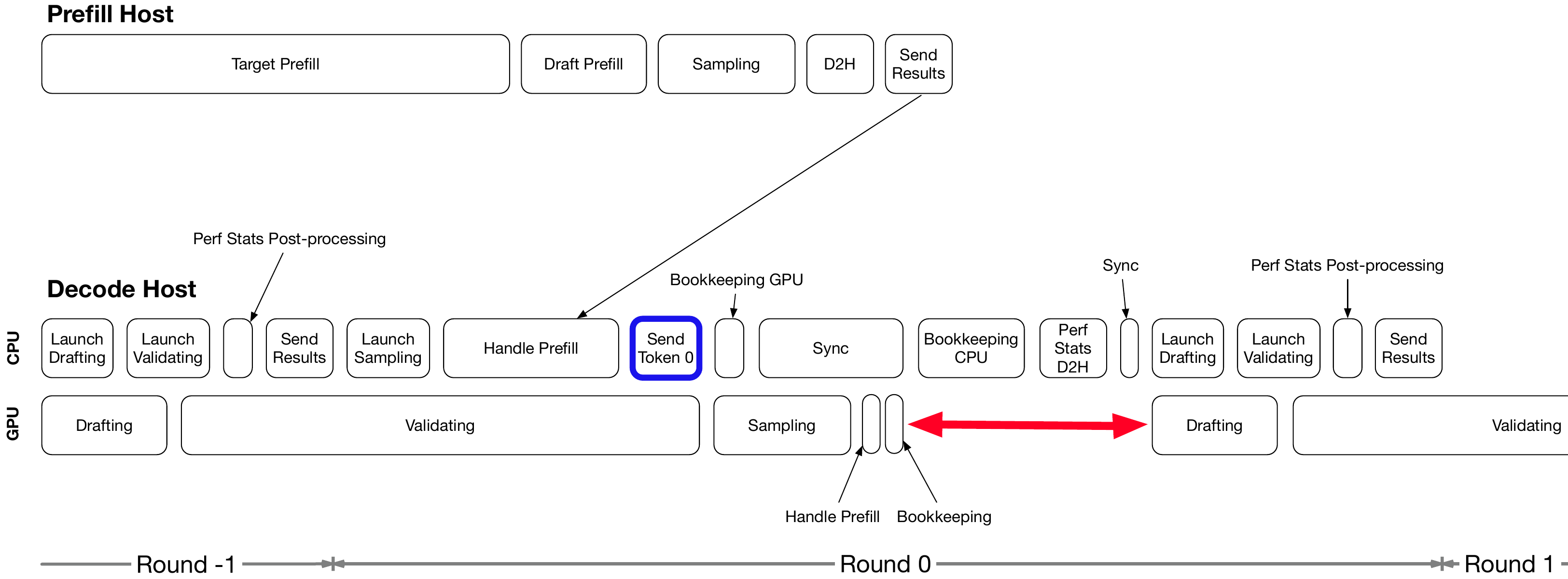}
        \caption{After latency hiding optimizations}
        \label{fig:hide_cpu_ops_trace_after}
    \end{subfigure}
    \caption{Disaggregated decoding cycle restructuring optimized inference efficiency by overlapping CPU and GPU tasks. This reduced GPU idle time and improved overall efficiency, as indicated by the \textcolor{red}{red arrows} in \textbf{(a)} and \textbf{(b)}.}
    \label{fig:hide_cpu_ops_trace}
\end{figure}

\subsection{Disaggregated inference with large batch sizes} 

In production environments, we use a disaggregated approach,  prefill and decode operations are performed on separate hosts  (see \Cref{fig:hide_cpu_ops_trace}). In this setup, the client communicates with the decode host, which redirects requests to prefill hosts for the first iteration and handles the remaining iterations itself. Due to this disaggregation, speculative decoding must accommodate large batch sizes and variable traffic, which can be challenging as it complicates efficient computational resource management. Furthermore, the increased demand for FLOPs for larger batch sizes requires careful allocation of hardware resources to ensure that the system can handle peak loads without bottlenecks, including out-of-memory errors. 

\begin{figure}[t!]
    \centering
    \begin{subfigure}[b]{\columnwidth}
        \centering
        \includegraphics[height=175px]{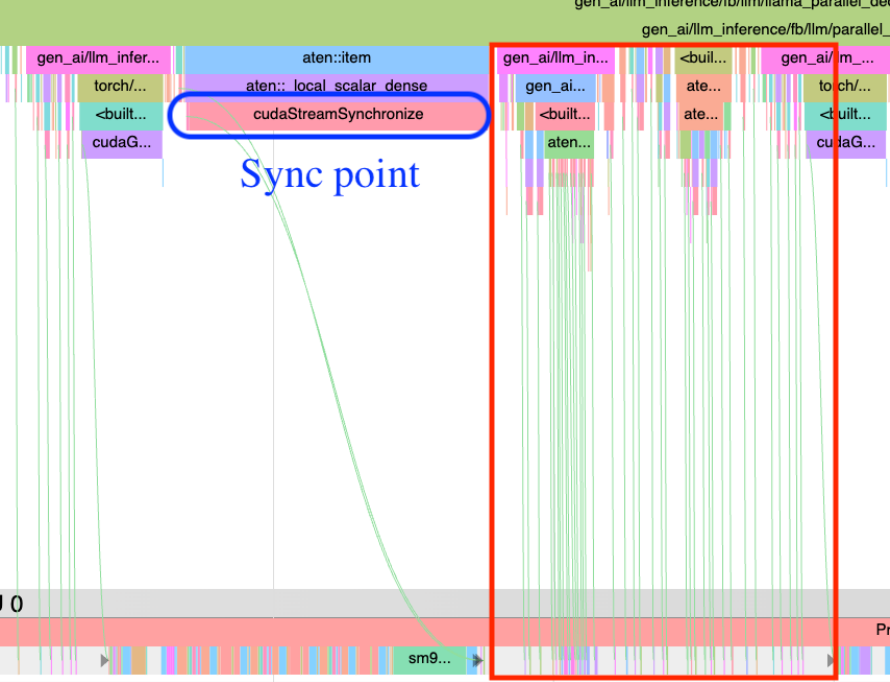}
        \caption{Before optimization}
        \label{fig:remove_cpu_gpu_sync_trace_before}
    \end{subfigure}
    \vfill
    \begin{subfigure}[b]{\columnwidth}
        \centering
        \includegraphics[width=0.85\columnwidth]{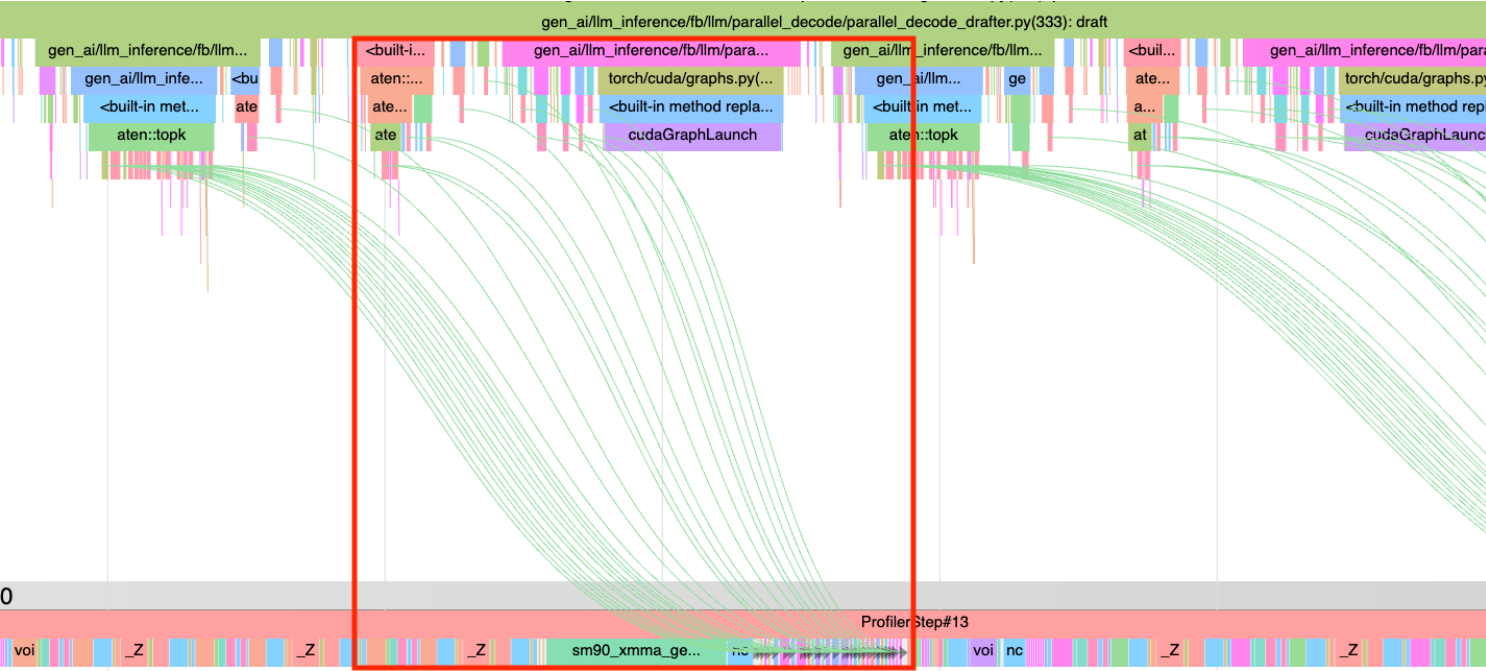}
        \caption{After optimization}
        \label{fig:remove_cpu_gpu_sync_trace_after}
    \end{subfigure}
    \caption{An example showing the effect of removing unnecessary CPU-GPU synchronization point from decoding flow. Initially, the CPU waited for GPU execution (highlighted as \textcolor{blue}{blue box} in \textbf{(a)}), causing slowdowns (highlighted as \textcolor{red}{red box} in \textbf{(a)}). After removing synchronization points, CPU kernels launched ahead of GPU execution, reducing idle time (highlighted as \textcolor{red}{red box} in \textbf{(b)}).
    }
    \label{fig:remove_cpu_gpu_sync_trace}
\end{figure}

To handle large batch-sizes with speculative decoding, we implemented the following optimizations that improve execution efficiency and GPU utilization to 94\%:
\begin{itemize}
    \item We collected hardware traces to identify unnecessary CPU-GPU synchronization points and removed them. This allows the CPU to run ahead of the GPU, ensuring that the GPU has sufficient tasks in its work queue. This helps reduce GPU idle time and improve latency. An example trace is shown in \Cref{fig:remove_cpu_gpu_sync_trace}. At the synchronization point highlighted by the \textcolor{blue}{blue box} in \Cref{fig:remove_cpu_gpu_sync_trace_before}, the CPU waited for GPU execution to finish, resulting in inefficiencies and slowdowns in subsequent operations due to CPU kernel launch overhead and GPU idleness, as indicated by the \textcolor{red}{red box} in \Cref{fig:remove_cpu_gpu_sync_trace_before}. After removing this synchronization point, as shown in \Cref{fig:remove_cpu_gpu_sync_trace_after}, CPU kernels were launched well ahead of GPU execution. This optimization eliminated GPU idle time and improved TTIT by 0.4ms. On an average, we found that TTIT was improved by 8-12\% after removing all possible CPU-GPU synchronization points.
    
    \item We restructured the decoding cycle by overlapping CPU operations with GPU kernel execution to further improve latency. Specifically, we decomposed tasks (e.g.,  bookkeeping) into distinct GPU and CPU components. This allowed us to concurrently execute CPU post-processing tasks while GPU kernels are running, effectively hiding CPU operations behind GPU processing and improving overall system efficiency. As shown in \Cref{fig:hide_cpu_ops_trace}, there are gaps between bookkeeping, prefill handling, and drafting, as highlighted by \textcolor{red}{red arrows} in \Cref{fig:hide_cpu_ops_trace_before}. With restructuring, we were able to reduce the GPU idle time (see \textcolor{red}{red arrow} in \Cref{fig:hide_cpu_ops_trace_after}), which helped in improving the TTIT on an average by about 10\%.
    
    \item To optimize the time to first token (TTFT), we added a sampling step at the end of the prefill phase (indicated by the \textcolor{blue}{blue box} in \Cref{fig:hide_cpu_ops_trace_after}). This allows us to immediately consume the outputs (hidden state and next predicted token) of the first token as soon as they are received by the decoder, rather than waiting for the entire decoding cycle to complete. By doing so, we reduce latency and improve response time by approximately 8-30\% on average, depending on the traffic volume.
    \item We reordered the processing sequence by moving prefill response handling ahead of bookkeeping, allowing us to hide it behind the long-running validation kernels. Additionally, we initiated the subsequent round of drafting and validating kernels earlier, effectively masking the latency associated with transmitting results to the client.
\end{itemize}

\subsection{Other performance optimizations}

\begin{figure}[t!]
    \centering
    \begin{subfigure}[b]{\columnwidth}
        \centering
        \begin{tabular}{cc}
            \includegraphics[width=0.48\columnwidth]{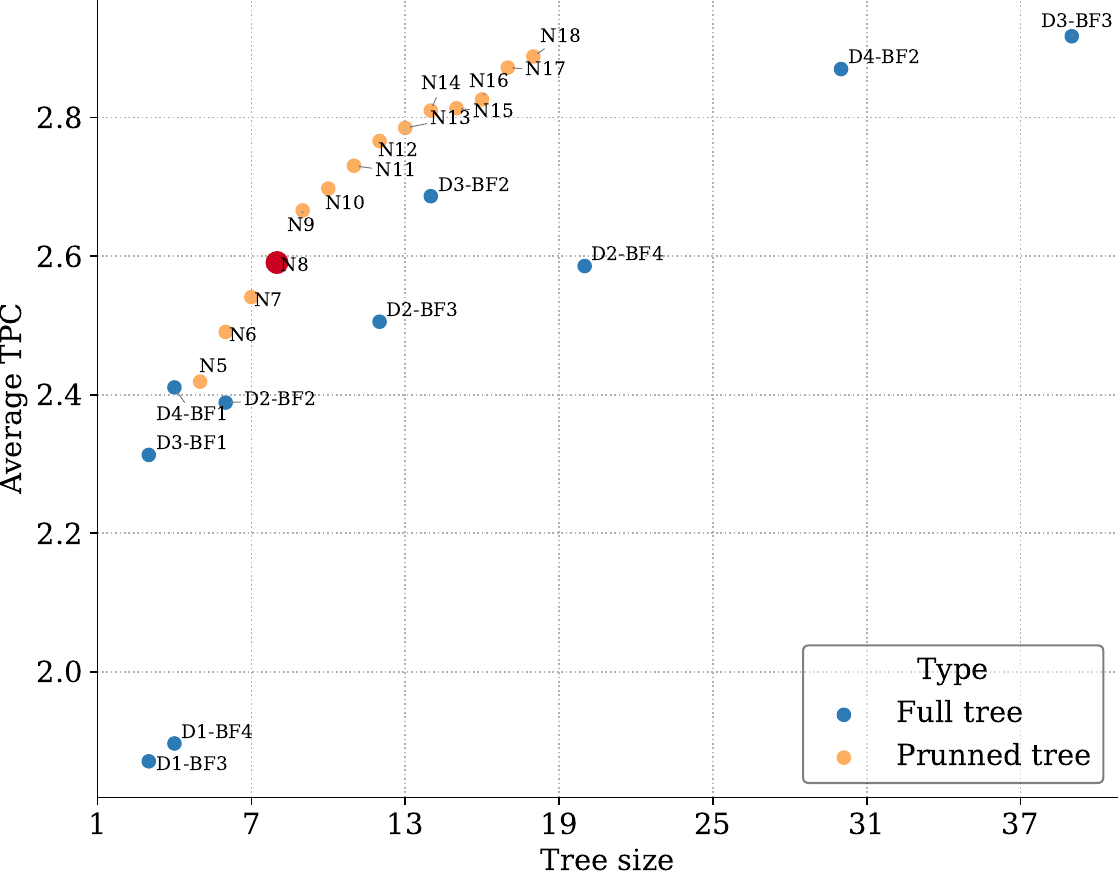} & \includegraphics[width=0.48\columnwidth]{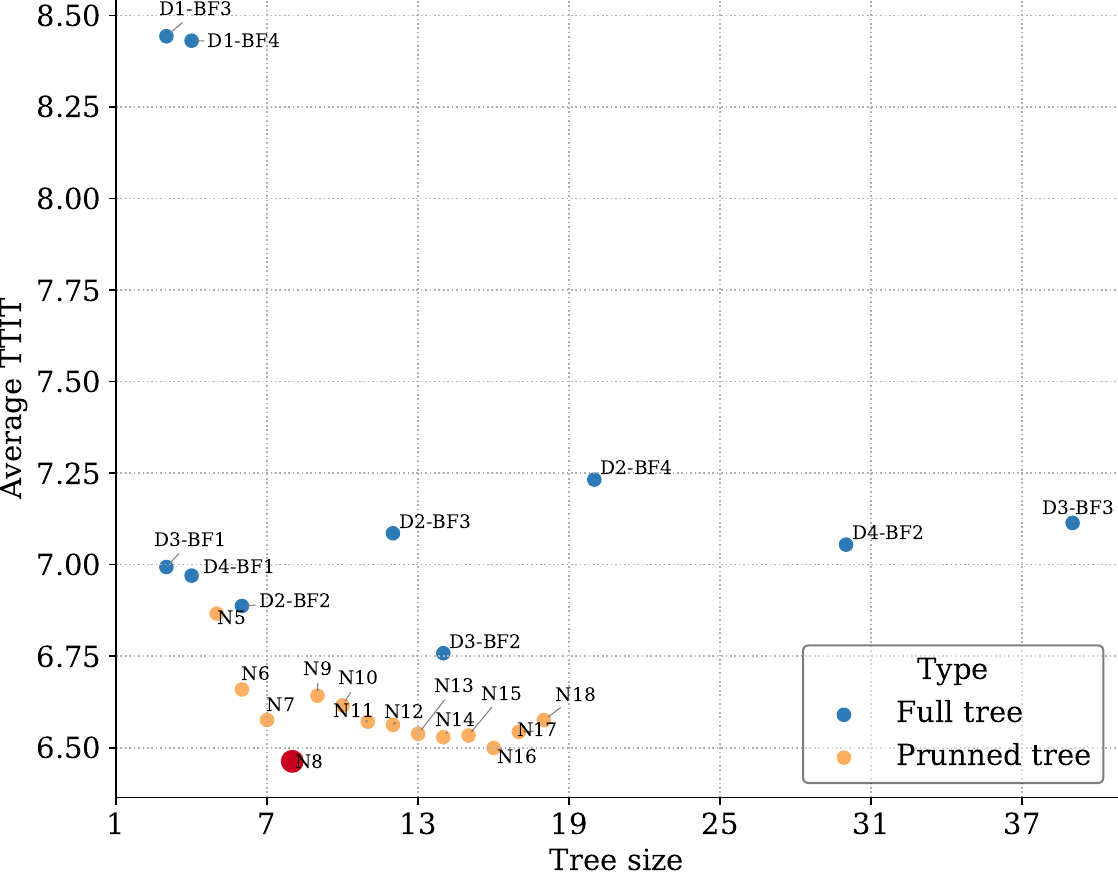} \\
        \end{tabular}
        \caption{Llama3.3 70B (Dense)}
    \end{subfigure}
    \hfill
    \begin{subfigure}[b]{\columnwidth}
        \centering
        \begin{tabular}{cc}
            \includegraphics[width=0.48\columnwidth]{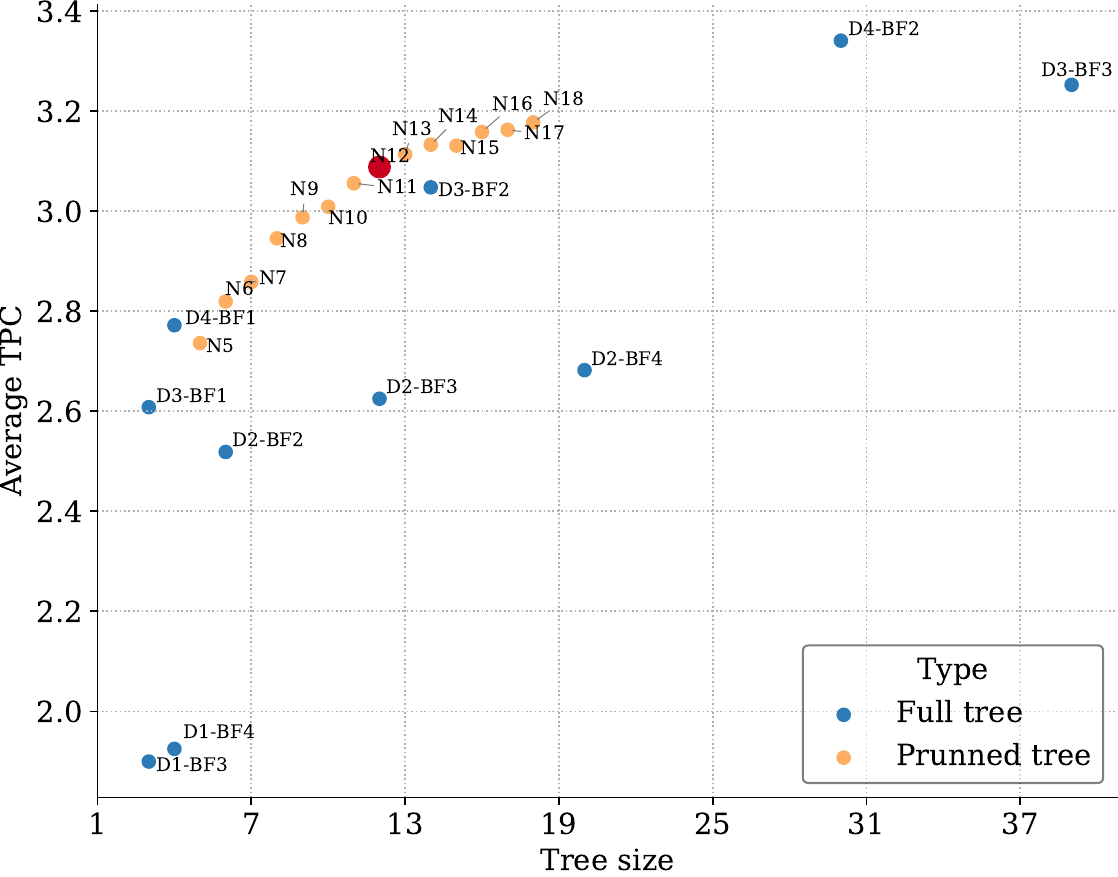} & \includegraphics[width=0.48\columnwidth]{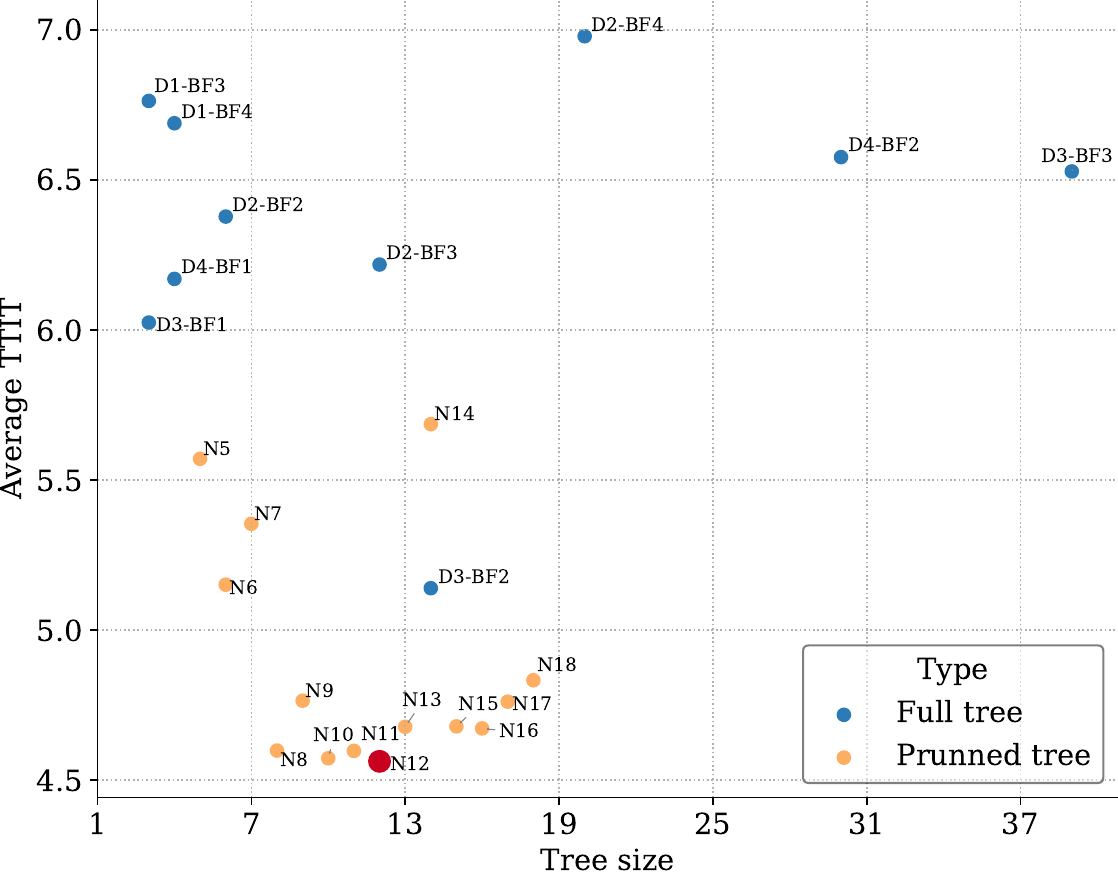} \\
        \end{tabular}
        \caption{Llama4 Scout (MoE)}
    \end{subfigure}
    \caption{Effect of different tree structures on TPC and TTIT for two different Llama models. We study two different tree structures: (1) a \textcolor[HTML]{2c7bb6}{\bfseries full tree} with depth $m$ and branching factor $n$, and is represented as $\text{D}_{m}\text{-BF}_{n}$ in graph, and (2) \textcolor[HTML]{fdae61}{\bfseries prunned tree} with $i$ nodes, and is represented as $N_{i}$ in a graph. The optimal tree configurations that are used in our production environments for both models are different and highlighted in \textcolor[HTML]{ca0020}{\bfseries red} (N8 and N12 for Llama3 70B and Llama4 Scout respectively). Here, TTIT is measured with a batch size of one on NVIDIA H100 GPUs. Here, TPC is  measured with temperature=0, top-p=0.9, and speculation length of three.}
    \label{fig:tree_ttit_tpc}
\end{figure}

\paragraph{Pre-computed static trees.} In our production environment, the batch size varies widely, ranging from 1 to over 100. This variability presents a challenge, as no single tree structure is optimal across the entire range. \Cref{fig:tree_ttit_tpc} shows that larger tree structures (which are effective at small batch sizes) yield higher TPC but are slow. Furthermore, at larger batch sizes, the computational cost outweighs the benefits of increased TPC. To select the optimal tree configurations, we developed \emph{tree dispatcher}. Based on a pre-computed static tree structures, it decides which tree configuration to use for the current batch size. For example, for a model typically running at small batch sizes we can decide to use a static tree when batch size is 1 and a chain draft of 3 tokens when batch size is larger than 1. This way we account for a trade-off between the computational cost and TPC, and ensure optimal performance across all traffic conditions.

\paragraph{Draft KV cache alignment} At the beginning of each drafting stage, the EAGLE method overwrites the draft model's KV cache by executing a forward pass using the previous round's newly accepted tokens and hidden states. Synchronizing the KV cache at this stage ensures that the draft model incorporates relevant contextual information from the previous validation round, resulting in more accurate and coherent drafting.  However, this approach incurs a non-negligible additional computation cost. We found that using only the last output token and its corresponding hidden states from the previous round is sufficient to align the KV cache of the draft model with the base model while improving inference efficiency. 

\paragraph {CUDA graphs.} During inference, particularly with large batch sizes, the drafting and validation stages are the most computationally demanding due to model execution. We expand the usage of CUDA Graphs in Meta's internal inference engine to optimize drafting and validation. This effectively eliminates kernel launch overhead and streamlines GPU operations. Notably, we chose to capture the entire model execution in a CUDA graph for maximum performance. This approach contrasts with some open-source inference libraries (e.g., vLLM-v1 \citep{kwon2023efficient}), which separate attention computation from CUDA graph capture to offer greater flexibility.

\paragraph{Attention kernel} Attention used in the decoding process of LLMs is bottlenecked by the need to load keys and values from the cache located in GPU high-bandwidth memory (HBM). As a result, the runtime scales linearly with the context length. There has been several efforts to implement attention efficiently on a given hardware \citep[e.g.,][]{flashdecoding2023, dao2022flashattention, dao2023flashattention2}. 

For faster decoding in production environments, we use  hardware-specific attention kernels. In our environment, we have three options: (a) Flash Decoding (or Triton Split-k kernel), (b) Flash Attention v2 and v3, and (c) Composable Kernel on AMD  \citep{composable_kernel}. Based on the batch size, tree size, and GPU type, we use a heuristic-based method to select among these attention options, which is primarily derived from the KV split strategy in these approaches. For example, the flash decoding kernel outperforms the others in terms of latency when the tree size is $\le 4$ and the batch size is $\le 64$. Flash attention v3 provides the best latency for larger trees and batch sizes when supported by the hardware.

\paragraph{Paged KV with tree attention} Paged KV partitions the KV cache into blocks and allocate the blocks as needed \citep{kwon2023efficient}. We used paged KV to optimize memory usage, and maintain separate caches for base and draft models. However, paged KV is not compatible with tree attention. To make it compatible, we implemented two changes: (a) the suffix part in tree attention is relatively small. To minimize the overhead of looking up paged blocks, we only apply paging to the prefix attention, and write back the intermediate keys and values instead of retrieving them from the cache, and (b) we add additional tree padding when allocating blocks for drafting and validation stages to prevent speculative tokens from exceeding block boundaries.

\paragraph{Persistent KV} Our system leverages persistent KV caching to enhance cross-request performance. As requests are processed, least recently used (LRU) blocks from the paged KV cache are evicted to the persistent KV cache in order to accommodate new KVs. Then, during prefill, these cached KVs are reused for matching prompt prefixes, thereby reducing time to first token (TTFT).

We maintain separate persistent KV caches for base and draft models, ensuring that both can benefit from this optimization. Additionally, our system enforces that both models show similar cache hit and block eviction behaviors for any given sequence of tokens. This was done primarily to simplify the implementation and improve maintainability. To achieve this uniform behavior, we made the following adjustments:
\begin{itemize}
\item We prefill both the base and draft model KV caches with the entire sequence of prompt tokens. However, for drafting, EAGLE requires the hidden state of the previous token from the base model, which is not available for the first prompt token. Unlike EAGLE, which skips the first prompt token during draft prefill, we fuse the first token with its own hidden state outputted from the base model prefill. Empirically, we have observed that this approach results in a slight increase in TPC.
 \item To maintain synchronization between the draft and base models, we scale the draft model’s KV cache size by a factor of $\frac{\text{number of draft layers}}{\text{number of base layers}}$ relative to the base model’s KV cache size. This ensures that both the paged and persistent caches of both models have identical sets of blocks, leading to synchronized block eviction times.
\end{itemize}

\begin{table}[t!]
    \centering
    \begin{tabular}{lcccccc}
        \toprule[1.5pt]
        & \multicolumn{3}{c}{\textbf{TPC} $\bm{\uparrow}$} & \multicolumn{3}{c}{\textbf{Drafting Latency (ms)} $\downarrow$}  \\
        \cmidrule[1.25pt](lr){2-4} \cmidrule[1.25pt](lr){5-7}
        & \textbf{BF16} & \textbf{FP8} & \textbf{INT4} & \textbf{BF16} & \textbf{FP8} & \textbf{INT4} \\
        \midrule[1pt]
        Llama3.1 8B (Dense) & 2.79 & 2.79 & 2.76 & 1.00 & 0.93 & 0.96 \\
        Llama3.3 70B (Dense) & 2.95 & 2.95 & 2.96 & 1.00 & 0.89 & 0.83 \\
        Llama4 Scout (MoE) & 2.86 & 2.86 & 2.87 & 1.00 & 0.89 & 0.87 \\
        Llama4 Maverick (MoE) & 2.81 & 2.79 & 2.78 & 1.00 & 0.93 & 0.92 \\
        \bottomrule[1.5pt]
    \end{tabular}
    \caption{Effect of FFN quantization in draft models. Here, TPC is measured with chain-like draft, temperature=0, and top-p=0.9 on MT-Bench. Also, drafting latency is the time to generate three tokens (corresponding to speculation length in our experiments) auto-regressively from the draft model.}
    \label{tab:int4_latency_metrics}
\end{table}

\paragraph{Quantized draft model} 
The ``losslessness" aspect of speculative decoding is that the draft model's quality does not affect the final output; it only impacts the speed. Therefore, we can compress the draft model differently than the base model. \Cref{tab:int4_latency_metrics} shows that INT4 feed-forward network quantization provides a good trade-off between TPC and decoding speed.

\paragraph{Guided decoding} Guided decoding helps generate structured outputs and is crucial in production environments \citep{willard2023efficient}. To enable speculative decoding for guided decoding requests, we implemented two changes: (a) We integrated guided decoding logic across all stages of the speculative decoding workflow, including drafting, sampling, and bookkeeping. (b) We improved performance by optimizing GD-related operations on the GPU. Specifically, we efficiently initialized the GD finite state machine (FSM) states and accelerated the key step of masking logits by moving CPU-related operations to the GPU. This avoided synchronization overhead, and we observed a speed-up of $2.6\times$ when guided decoding with speculative decoding was used compared to the non-speculative decoding baseline. We note that results presented in this paper are without guided decoding.

\paragraph{iRoPE for Llama4} Llama4 introduced interleaved ROPE (iRoPE), a variant of local attention where sequences are split into sub-sequences of fixed length, and queries can only attend to values within the same sub-sequence. This effectively modifies the attention mask, making the blocks in the block-diagonal mask smaller. Recall that tree attention (see  \Cref{ssec:tree_attention}) also modifies the attention mask. Therefore, it is essential to combine these two modifications correctly for Llama4 inference.

The iRoPE attention mask for decoding was implemented using XFormers' gappy attention biases. We adapted the tree attention to use XFormers' gappy attention biases in prefix attention and integrated iRoPE logic through the validation path. One corner case that we encountered was when the draft string crosses the boundary of two sub-sequences. While it is theoretically possible to implement complex logic for splitting the tree mask between two sub-sequences, we opted for a simpler solution: truncating the draft to fit entirely within one sub-sequence. As such situations are extremely rare, the impact on the acceptance rate is negligible.

\begin{figure}[t!]
    \centering
    \begin{subfigure}[b]{0.48\columnwidth}
        \includegraphics[width=0.925\columnwidth]{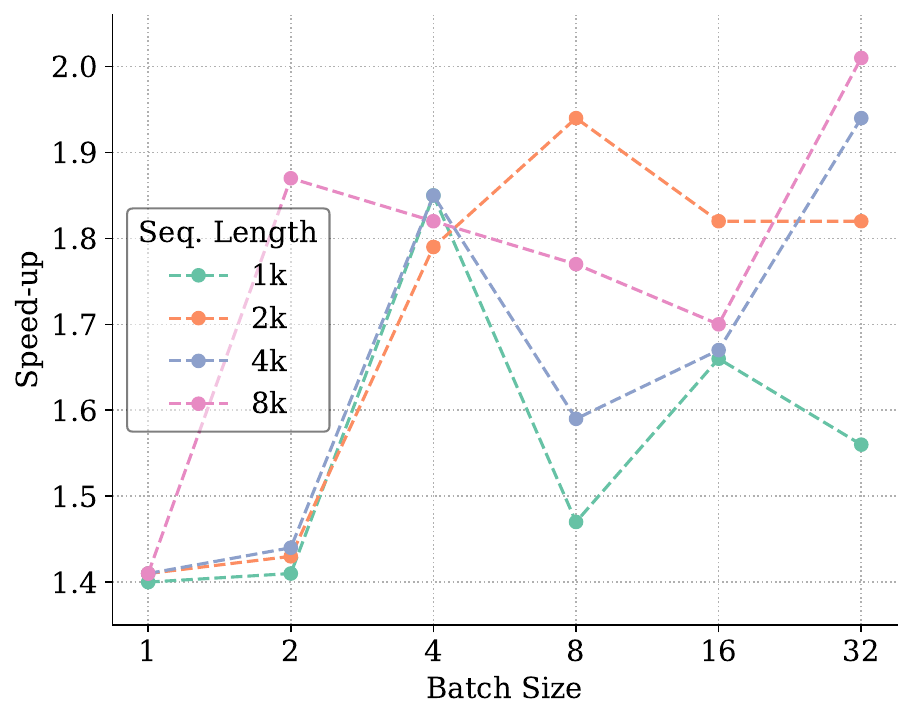}
        \caption{Llama3.1 8B (Dense)}
        \label{fig:l3_8b_speedup}
    \end{subfigure}
    \hfill
    \begin{subfigure}[b]{0.48\columnwidth}
        \includegraphics[width=0.925\columnwidth]{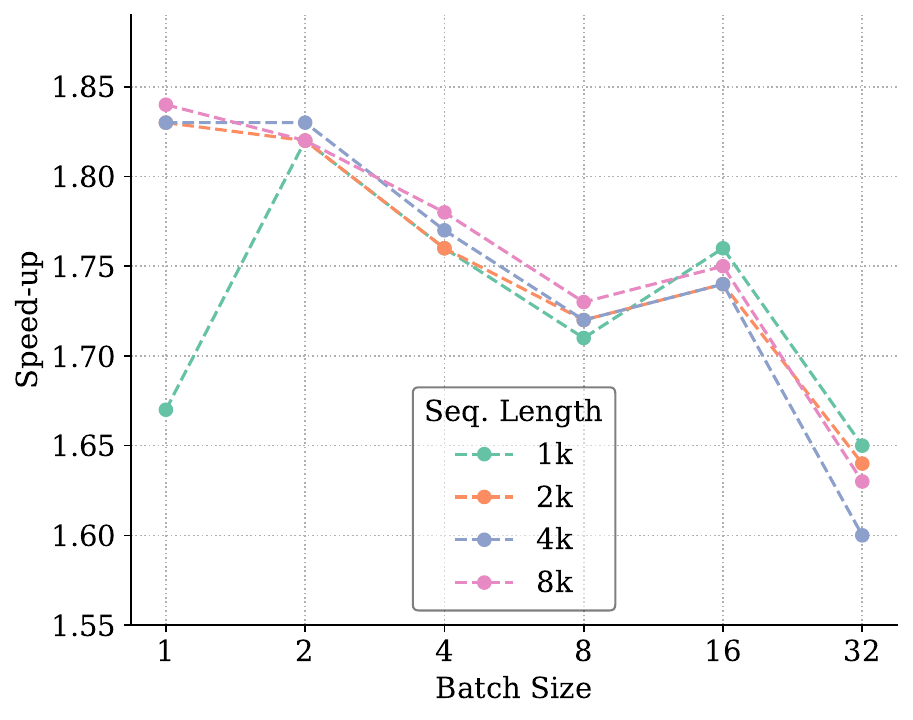}
        \caption{Llama3.1 70B (Dense)}
        \label{fig:l3_70b_speedup}
    \end{subfigure}
    \vfill
    \begin{subfigure}[b]{0.48\columnwidth}
        \includegraphics[width=0.925\columnwidth]{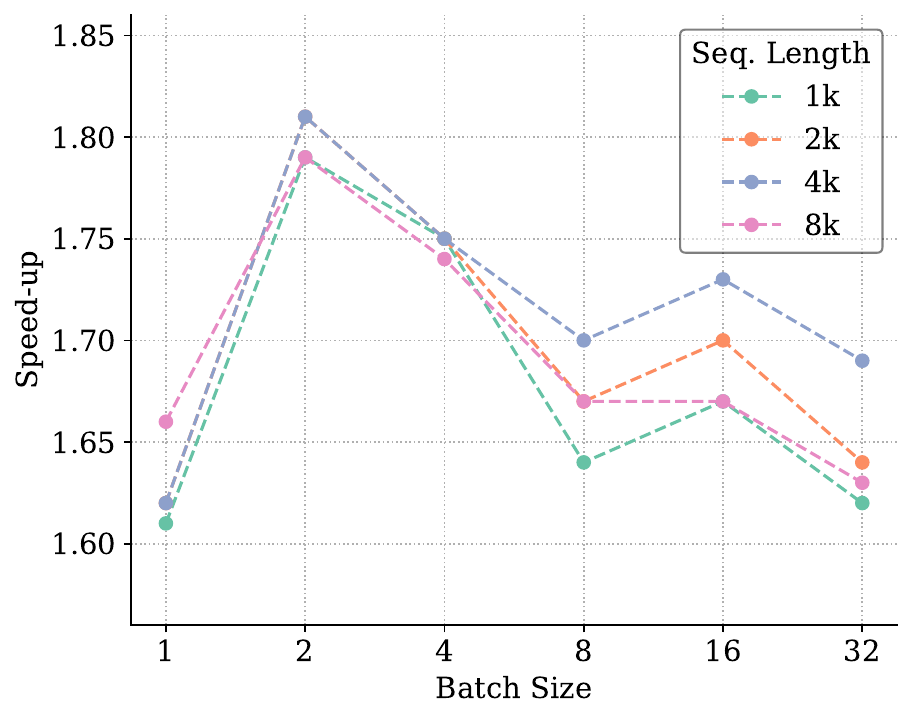}
        \caption{Llama4 Scout (MoE)}
        \label{fig:l4_scout_speedup}
    \end{subfigure}
    \hfill
    \begin{subfigure}[b]{0.48\columnwidth}
        \includegraphics[width=0.925\columnwidth]{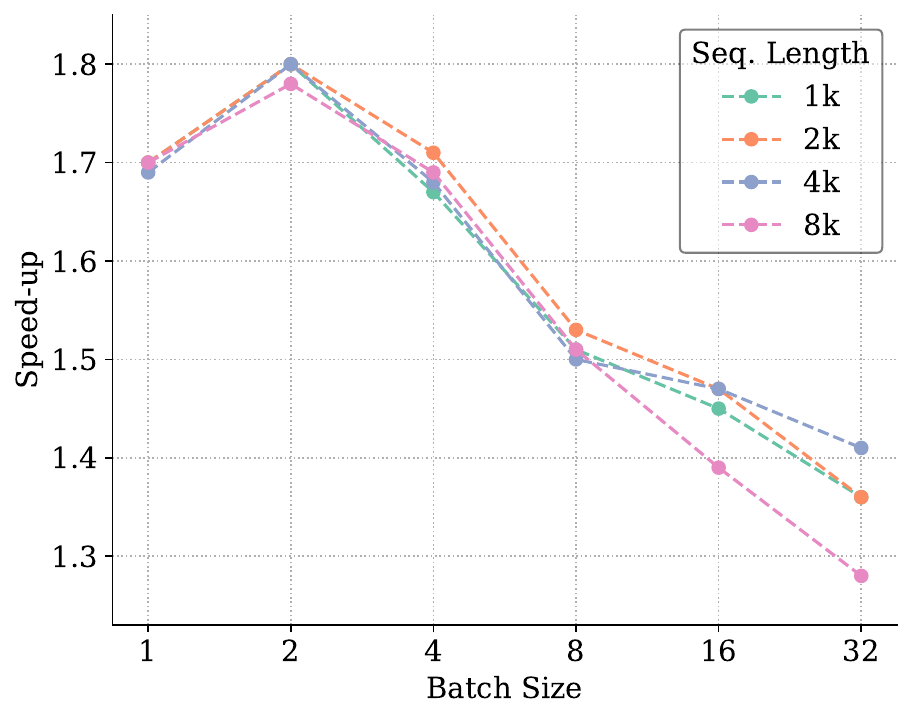}
        \caption{Llama4 Maverick (MoE)}
        \label{fig:l4_mav_speedup}
    \end{subfigure}
    \caption{Inference speed-up of various Llama models using EAGLE speculative decoding, measured relative to the baseline performance. The plot shows how speed-up varies with different batch sizes and sequence lengths.}
    \label{fig:inference_speed_up}
\end{figure}

\subsection{Benchmarking}

Many previous works assumed that speculative decoding speed-up decreases as batch size increases \citep{su2023synergyspeculativedecodingbatching, miao2024specinfer, liu2024optimizingspeculativedecodingserving}. This is likely because speculative decoding methods may be under-utilizing GPU \emph{computational resources}, i.e., increased FLOPs from running drafting and validation comes for free because decoding is bound by the GPU memory bandwidth and not compute. However, at large batch sizes, decoding becomes compute-bound, resulting in a decrease in speedup.

At large context lengths, attention dominates the computation even for large batch sizes, and therefore the workload stays memory-bound \citep{sadhukhan2025magicdec}. Our results in \Cref{fig:inference_speed_up} partially confirm these observations. Overall, we observe that the relationship between end-to-end speculative decoding speed-up and batch size varies depending on the model size and its performance characteristics. For instance, the Llama3.1 8B model exhibits greater speculative decoding speedup at large batch sizes compared to small batch sizes. In contrast, the speed-up for Llama4 Maverick, which has approximately 400 billion parameters, decreases with increasing batch size.

\section*{Contributors}

This project is the result of efforts by numerous individuals within the GenAI and Infra teams at Meta. Below, we acknowledge all core contributors and contributors, listed in alphabetical order by first name.

\textbf{Core contributors.} Bangsheng Tang, Carl Chengyan Fu, Fei Kou, Grigory Sizov, Haoci Zhang, Jason Park, Jiawen Liu, Jie You, Qirui Yang, Sachin Mehta, Shengyong Cai, Xiaodong Wang, Xingyu Liu, Yunlu Li, Yanjun Zhou, Wei Wei, Zhiwei Zhao, and Zixi Qi.

\textbf{Contributors.} Adolfo Victoria, Aya Ibrahim, Bram Wasti, Changkyu Kim, Daniel Haziza, Fei Sun, Giancarlo Delfin, Emily Guo$^\dagger$\footnote{$^\dagger$ Work done while working at Meta.}, Jialin Ouyang, Jaewon Lee, Jianyu Huang, Jeremy Reizenstein, Lu Fang, Quinn Zhu, Ria Verma$^\dagger$, Vlad Mihailescu, Xingwen Guo, Yan Cui, Ye Hu, and Yejin Lee.

\section*{Acknowledgments}
We thank Chuanhao Zhuge, Emad El-Haraty, Lai Wei, Mohammad Rastegari$^\dagger$, Rajasi Saha, Seiji Yamamoto, Sergey Edunov, Shaun Lindsay, Sijia Chen, and Tony Liu for their support. We also thank Edward Yang who helped to make \texttt{torch.compile} work well on multiround speculative sampling. We also thank the broader GenAI and Infra team members for the discussions and feedback.

\clearpage
\newpage
\bibliographystyle{assets/plainnat}
\bibliography{paper}

\end{document}